# Re-Training StyleGAN - A First Step Towards Building Large, Scalable Synthetic Facial Datasets

Viktor Varkarakis
Collage of Engineering and Informatics
National University of Ireland, Galway
Galway, Ireland
v.varkarakis1@nuigalway.ie

Shabab Bazrafkan
Imec VisionLab, Physics Department
University of Antwerp
Antwerp, Belgium
shabab.bazrafkan@uantwerpen.be

Peter Corcoran
Collage of Engineering and Informatics
National University of Ireland, Galway
Galway, Ireland
peter.corcoran@nuigalway.ie

*Abstract*—StyleGAN is a state-of-art generative adversarial network architecture that generates random 2D high-quality synthetic facial data samples. In this paper we recap the StyleGAN architecture and training methodology and present our experiences of retraining it on a number of alternative public datasets. Practical issues and challenges arising from the retraining process are discussed. Tests and validation results are presented and a comparative analysis of several different re-trained StyleGAN weightings is provided [1]. The role of this tool in building large, scalable datasets of synthetic facial data is also discussed.

## I. INTRODUCTION

In the last few years, a number of tools for generating synthetic facial data samples have evolved [1]–[3], based on generative adversarial networks (GANs) [4]. These enable photo-realistic, high-resolution random face samples to be generated in almost limitless numbers. However, these face samples are essentially random with no relationship to one another, and thus of limited value or interest to researchers.

Some research has attempted to modify the training of these GANs to provide more control over the output data samples. As examples, Bazrafkan et al [5] have shown that a GAN can be trained so that the first vector in the latent space represents male samples if it has a negative value or female images if the value is positive. In a separate work, Bazrafkan et al [6] have shown that faces can be generated in a fixed pose set by an auxiliary regressor.

From these works, it is clear that GANs can be adapted and modified to enable more sophisticated control of the generation of synthetic facial data samples. It is also important to understand the relationship between the original training dataset and the data samples obtained from the resulting GAN. In order to obtain a wide variation in output data, it is necessary to have a large original training dataset, but it is not clear from current research how the size and quality of this original dataset affect the distribution of the output data samples. Neither is it clear how well these synthetic data samples are distinguished from the original data.

StyleGAN has been used widely and trained on different image topics (cats, cars, bedrooms, anime, etc) [3][7]. The original models released in StyleGAN [3], provide models trained on the FFHQ [3] and CelebA-HQ [8] datasets at a resolution of 1024×1024. It is worth noting that the size of these datasets is relatively small as they consist of 70000 and 30000 samples respectively. Apart from these original models to the best of our knowledge a few models are publicly available that provide StyleGAN models trained on alternative facial datasets at different image resolutions and quality. These models are unofficial implementations [2],[3] of the StyleGAN in other frameworks (PyTorch) and therefore it is not possible to compare them with the original StyleGAN [3].

As a first step to understanding better the nature of state-of-art GANs we have re-trained StyleGAN on a number of large publicly available datasets and made the resulting model networks with their weights publicly available [1]. In this paper, we describe the re-training process on various original facial datasets and explain the various steps in the re-training process. We then perform a comparative analysis of a randomized set of data created by each of the re-trained GANs, in order to evaluate the quality and diversity of the generated samples.

## II. AN OVERVIEW OF PUBLICLY AVAIALBLE FACIAL DATASETS

In recent years more and more "in the wild" face datasets have become available of different sizes and with each dataset being proposed for a different use case. A short description of some of the more significant publicly available facial datasets is given below.

### A. LFW

Labelled Faces in the Wild (LFW) [9] is the de facto standard test dataset for the face verification in unconstrained conditions. The majority of research publications related to the face verification task report their performance with the mean face verification accuracy and the ROC curve on the standard evaluation set of 6,000 given face pairs in LFW. The dataset was released in 2007 and contains 13,233 face images of 5,749 identities. Although, it should be mentioned that due to the small number of identities and the number of samples per identity in the LFW, it is inadequate for training purposes and thus is used mainly for testing.

---

[1] https://github.com/C3Imaging/Deep-Learning-Techniques/tree/Re-training-StyleGAN
[2] https://github.com/kayoyin/FakeCelebs
[3] https://github.com/podgorskiy/StyleGan

### B. CASIA-WebFace

CASIA-WebFace [10] is one of the first large public facial datasets, published in 2014. It contains 10,575 identities with a total of 494,414 facial data samples. The identities belong to celebrities and all of them are collected from the IMDb website. The size of the dataset makes it suitable for facial recognition tasks and this dataset is frequently used as a baseline by researchers in the facial recognition filed.

### C. CelebFaces

The CelebFaces+ dataset [11] was released in 2014 and along with the CASIA-WebFace was one of the first large publicly available datasets, as it contains 202,599 images of 10,177 identities. A version of this dataset with additional metadata is known as CelebFaces Attributes Dataset (CelebA) [12] where the samples form CelebFaces+ are annotated with 5 landmark locations and for 40 binary attributes (eyeglasses, moustache, hat, etc), providing valuable information for the researchers.

### D. MegaFace

The MegaFace dataset [13] was published in 2016 in order to examine face recognition methods with up to a million distractors in the gallery image set. The dataset consists of 4.7M samples organised into 672,057 identities. Despite being a large dataset, it offers a limited set of variations per identity, as on average it has only 7 samples per person. Depending on the specific research goal this can limit the usefulness of MegaFace for some research tasks.

### E. Ms-Celeb-1M

The Ms-Celeb-1M dataset [14] was created and published in 2016 by Microsoft. It is the largest publicly available face recognition dataset with over 10M samples from 100K identities. The dataset is suitable for both training and testing purposes with an average of 100 samples per identity.

### F. VGGFace & VGGFace2

The VGG datasets are released from the Visual Geometry Group from the University of Oxford. The VGGFace [15] dataset was released in 2015 and contains 2.6M samples from 2,622 people. VGGFace was released mainly for training purposes. In 2018, the VGGFace2 [16] was released. This dataset comprises 3.31M samples from 9,131 celebrities – on average 360 samples per identity. The images were downloaded from Google Image Search. The image samples from VGGFace2 cover a wider range of different ethnicities, professions, and ages compared to VGGFace. Furthermore, all the samples have been captured "in the wild" thus giving the dataset a desirable variation with respect to pose, lighting and occlusion conditions as well as facial expressions. The dataset can be used for training and testing purposes as it is divided into a train and test set. Finally, VGGFace2 provides annotations regarding the pose and the age of its samples which can be useful for researchers.

### G. Other Face Datasets

Several other datasets should be mentioned such as YTF [17], which has 1,595 identities and 3,425 video clips. Another dataset that was built in order to recognise faces in unconstrained videos is UMDFaces-Videos [18], which consists of 3,107 identities and 22,075 video clips. Also, they have a face dataset as well with still images, the UMDFaces which consist of 367,88 samples from 8,277 identities [19].

Furthermore, FFHQ [3] and CelebA-HQ [8] are some face datasets that are not created for face recognition purposes. These datasets are of a high quality and a high resolution 1024×1024 compared to the aforementioned databases. These datasets were used to train StyleGAN, which produces high quality generated images. The FFHQ consists of 70,000 images without an identity annotation but contains variation in terms of age, ethnicity and image background. It also has a good coverage of accessories. CelebA-HQ is a subset of CelebA from 6,217 identities. As mentioned before the samples are at 1024×1024 resolution and of high quality which was achieved through a procedure of pre-processing that is explained in [8].

Finally, it is also worth remarking that large corporations, e.g. Facebook, Google, have their own in-house datasets that are likely to dwarf those that are publicly available. Facebook [20] trained some of their model with a dataset that comprises 500M facial samples from more than 10M identities and Google's model [21] was trained on 200M images from 8 million subjects.

### III. RETRAINING METHODOLOGY

In this section, the process of retraining StyleGAN is described. Initially, a preparatory filtering procedure is used to select the samples is explained and subsequently, the procedure of training StyleGAN is analysed. For the purposes of this work, the StyleGAN network was retrained twice, once with samples from the CelebA dataset and the other one is trained on the CASIA-WebFace dataset. The techniques documented here are being applied to additional datasets and corresponding results will be presented at the ISSC conference later this year.

### A. Data Sample Resolution and Quality Considerations

The datasets used in our experiments (CelebA, CASIA-WebFace) are pre-filtered before the data samples are fed into the training network. The filtering is performed for two main reasons. Firstly, it is important to ensure that the data samples given to the network contain a detectable face region of good quality. Most large face datasets contain noisy, poor samples and it is desirable to remove such samples as inputs to the training network as they can interfere with the main learning task of generating realistic face samples of good visual quality. Secondly, it is important to resize the facial samples to a particular size that the StyleGAN network will be trained on. The size of the image samples used in this work was determined to be 256×256 pixels. This offers a good balance between facial image quality and the computational resources required for training. With such a size of image samples, it is practical to train on a single dual-GPUs computer. Based on the information available on StyleGAN's GitHub repository [4], training at a higher resolution such as 1024×1024 with only two GPUs it would have required almost a month of continuous operation.

For our purposes – to gain practical experience in re-training these powerful GANs - samples at a resolution of 256×256 have sufficient visual information and quality for most practical uses and applications of synthetic facial data. It is also worth noting that most of the available public face datasets the data samples are of similar or lower resolution and often there is a significant number of samples that are noisy or of quite poor visual quality.

*B. Dataset Preparation & Pre-Filtering*

In the initial filtering step, the image samples are passed through a face detector. The face detector used is the Multi-task Cascaded Convolutional Networks (MTCNN) [22]. An implementation of the MTCNN in Python / TensorFlow was used which can be found in [5]. The face detector is applied to the images. The MTCNN implementation used takes two arguments, the margin size and the size of the output image. The margin used is 50px and the size of the image as mentioned earlier is 256×256. The CelebA dataset has 202,599 samples and the CASIA-WebFace has 494,414 samples. After the pre-processing procedure, CelebA and CASIA-WebFace consist of 202,281 and 491,073 samples respectively. In Fig 1. and Fig 2. some samples from CASIA-WebFace and CelebA are presented. The MTCNN detection was not able to confirm these samples as faces and thus they are not used in training. These examples illustrate the need for a pre-filtering step for the input data. In all large public face datasets, a significant number of such noisy data samples are expected and may unduly affect the training outcome. In Fig. 1 there are some examples of extreme pose, images with artifacts, blurred or extremely dark facial images or only partial face samples. In Fig. 2, samples are presented that only contain noise, or the face is mostly obscured and is not representative of a normal human face.

With the use of MTCNN to pre-filter the data, it is possible to a certain degree to eliminate many unwanted samples that would not be beneficial for retraining StyleGAN for the task of generating unobscured, human faces. Finally, in Fig. 3 and Fig. 4 a selection of good, high-quality, facial samples from CASIA-WebFace and CelebA are presented after the pre-processing procedure and used to prepare data samples for the main training procedure.

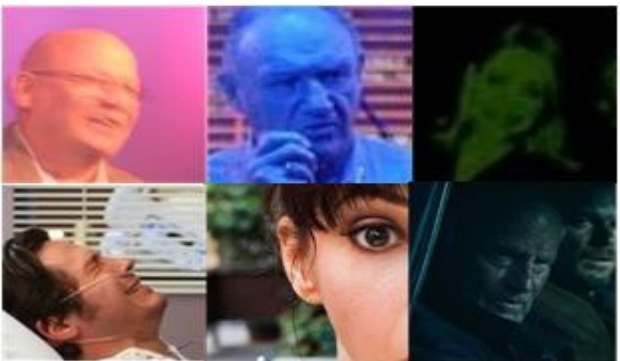

Fig. 1. CASIA-WebFace samples [10], not detected by the MTCNN [22].

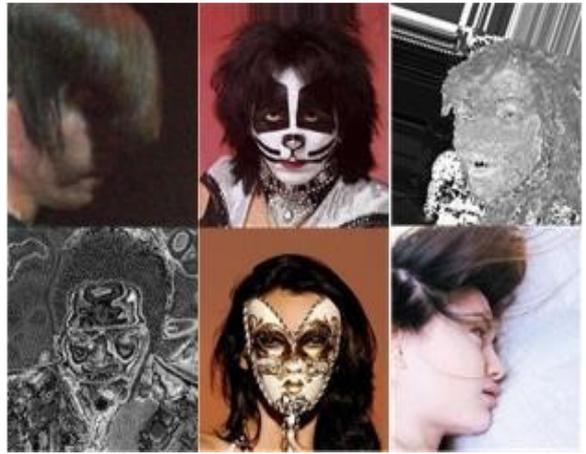

Fig. 2. CelebA samples [12], not detected by the MTCNN [22].

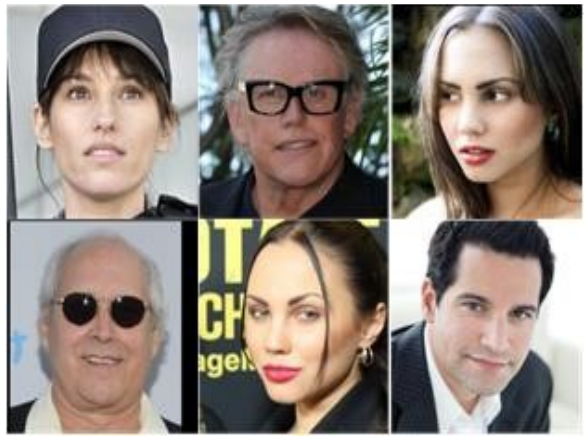

Fig. 3. CASIA-WebFace samples [10], after the pre-processing procedure

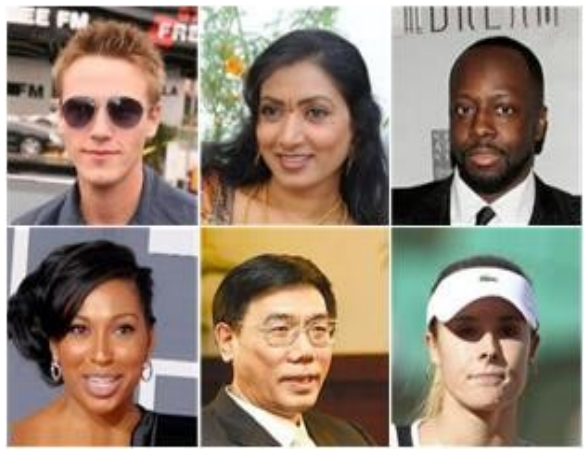

Fig. 4 CelebA samples [12] after the pre-processing procedure.

*C. The Re-Training Process*

The original Generative Adversarial Networks (GAN) presented in [4] is made of two Deep Neural Networks: a generator and a discriminator. The generator accepts a tensor of randomly generated numbers and returns an image and the discriminator is a binary classifier that accepts an image and determines whether it is a generated image or not. In this approach, these two networks are trained in a min-max game wherein the final goal is for the generator to synthesis an image that the discriminator classifies as a real image.

---

[4] https://github.com/NVlabs/stylegan

[5] https://github.com/davidsandberg/facenet

The StyleGAN [3] is one of the variations of GAN wherein the generator is developed in a specific way which separates it from its preceding implementations in three main ways:

1- The latent space (Z) is reshaped via a fully connected DNN (which returns W) before feeding into the generator. This is to introduce disentanglement to the original latent space (Z) during the mapping into style indicators (W).

2- The latent space is not fed into the generator at its input layer. The new latent space W is given to the generator before each convolutional layer. In other words, each part of the vector W is induced into the generator in a different layer. This gives the opportunity to introduce style information at different levels.

3- A Gaussian noise is added to the features before each convolution. This operation helps the network to use its maximum capacity and generate higher quality outputs with high-frequency features.

More details regarding the architecture as well as the hyperparameter selection of StyleGAN can be found in [3].

In this current study, once the samples were pre-processed, StyleGAN is re-trained on the two large datasets mentioned above (CelebA, CASIA-WebFace). The official implementation of StyleGAN was adopted to retrain on our databases. This implementation is in TensorFlow, requiring version 1.10 or newer and Python 3.6 and can be found in [4]. The default configuration for training was utilized which used to train the highest-quality StyleGAN with the FFHQ dataset at a 1024×1024 resolution. Full details can be found in [3], as here we only described modifications to the base training. As mentioned earlier a dual-GPU machine was used to train StyleGAN. The GPUs used are RTX 2080 Ti. The training process was performed twice, once training with samples from CelebA dataset and once with samples from the CASIA-WebFace dataset. Each experiment ran for 12 days. After the end of the training process, the epoch/checkpoint with the best Fréchet Inception Distance [23] using 50,000 images (FID50k) is selected. FID50k is an evaluation metric used in the training procedure. In the next section, examples and the evaluation results for the best models are presented. Finally, we make these models publicly available and can be found in [1].

IV. RETRAINING – EXPERIMENTS & VALIDATION

As mentioned, the model with the best Fréchet inception distance (FID) using 50,000 images for each dataset with which the StyleGAN was trained (CelebA, CASIA-WebFace) is selected as the final model. The FID is an evaluation metric that captures the similarity of generated images to real ones better than the Inception Score [23]. A lower FID score means better image quality and diversity of the generated images. For the best model selected for each dataset, the results of several quality and disentanglement metrics are presented along with visual examples. More information about the metrics and the way that are calculated can be found in [3].

Table I. Quality and disentanglement metrics for several StyleGAN models trained on different datasets and resolutions. Specifically, StyleGAN trained on CelebA and Casia-WebFaces at 256 ×256 and the original StyleGAN [3] trained on FFHQ at 1024×1024.

| Metric | Results | | | Description |
|---|---|---|---|---|
| | StyleGAN on CelebA | StyleGAN on CASIA-WebFace | StyleGAN on FFHQ [3] | |
| FID50k | 4.7842 | 4.5992 | 4.4159 | Fréchet Inception Distance using 50,000 images. |
| ppl_zfull | 191.9051 | 258.4270 | 664.8854 | Perceptual Path Length for full paths in $Z$. |
| ppl_wfull | 68.6066 | 81.7605 | 233.3059 | Perceptual Path Length for full paths in $W$. |
| ppl_zend | 190.5838 | 259.5282 | 666.1057 | Perceptual Path Length for path endpoints in $Z$. |
| ppl_wend | 56.4555 | 74.2621 | 197.2266 | Perceptual Path Length for path endpoints in $W$ |
| ls in Z | 143.2236 | 109.7136 | 165.0106 | Linear Separability in $Z$ |
| ls in W | 2.5235 | 3.1748 | 3.7447 | Linear Separability in $W$ |

*A. Evaluation of the re-trained StyleGAN models*

In Table I, the quality and disentanglement metrics for the best StyleGAN models on each dataset are presented. The selected model of StyleGAN trained on the CelebA dataset, has a better score in the perceptual path length for full paths and for the endpoints path in W and Z latent space, compared to the selected model trained on CASIA-WebFace. Perceptual path length [24] measures the difference between consecutive images (their VGG16 embeddings) when interpolating between two random inputs. Drastic changes mean that multiple features have changed together and that they might be entangled, therefore showing that the model trained with the CelebA dataset generates samples that its features are less connected between them compared to the ones from the model trained with CASIA-WebFace. The metric of linear separability shows the ability to classify inputs into binary classes. The better the classification the more separable the features. In this metric, the model trained with CelebA has a better score in the W latent space whereas the model trained on CASIA-WebFace has a better score in the Z latent space. Finally, in the FID score using 50,000 images the StyleGAN model trained on CASIA-WebFace has a better score illustrating a slightly better quality and diversity in the generated samples compared to the StyleGAN model from CelebA. In Table I, the quality and disentanglement metrics for the StyleGAN trained on FFHQ from [3], are presented. We do not compare with their results as they trained StyleGAN on high quality and high resolution (1024×1024) in contrast with our models which they were trained on a ×4 smaller resolution and lower quality samples.

Below several samples are generated from the trained StyleGAN models. Fig. 5 and 6 show an uncurated set of novel images generated from the trained generator. Fig. 7. shows the effect of applying stochastic variation to different subsets of layers. Fig 8 illustrates the effect of the truncation trick as a function of style scale ψ. The samples in Fig 5, and 8 are generated by the selected StyleGAN model trained with CelebA and Fig 6 and 7 with the selected StyleGAN model trained with CASIA-WebFace. The models used to generate the samples are available in [1].

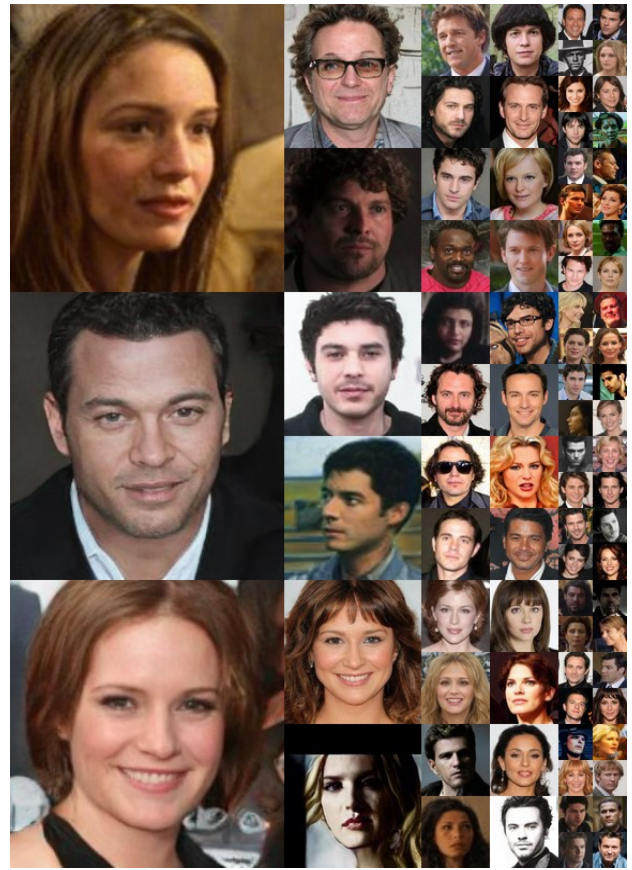

Fig. 6. Uncurated set of images produced by the best StyleGAN model trained with CASIA-WebFace. The samples are generated with a variation of the truncation trick [25]–[27], ψ = 0.7 for resolutions $4^2 - 32^2$. This figure is similar to the figure (2) from [3].

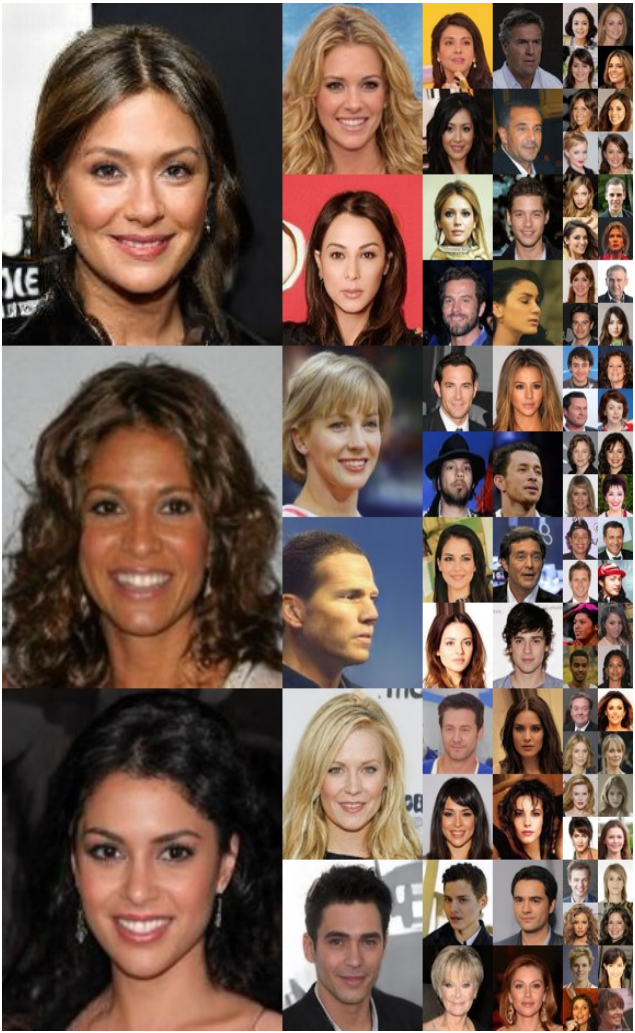

Fig. 5. Uncurated set of images produced by the best StyleGAN model trained with CelebA. The samples are generated with a variation of the truncation trick [25]–[27], ψ = 0.7 for resolutions $4^2 - 32^2$. This figure is similar to the figure (2) from [3]

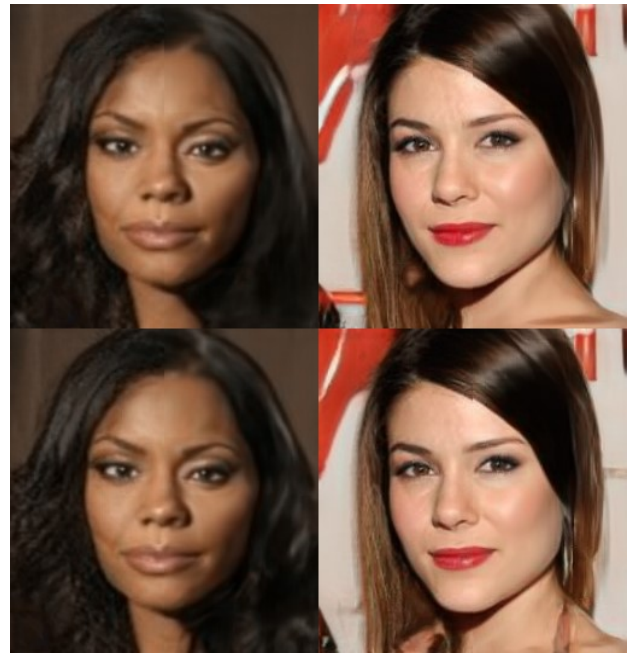

Fig. 7. Effect of noise inputs at different layers of the generator. (a) Noise is applied to all layers. (b) No noise. (c) Noise in fine layers only ($64^2 - 1024^2$). (d) Noise in coarse layers only ($4^2 - 32^2$). This is similar to figure (5) from [3].

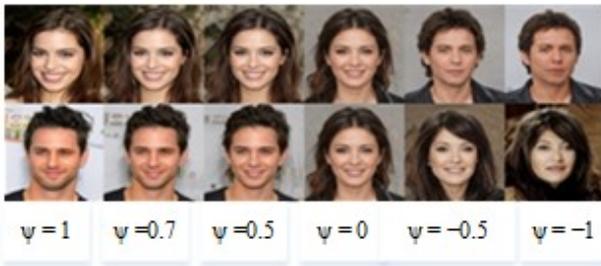

Figure 8 The effect of truncation trick as a function of style scale ψ (ψ=1). When we fade ψ → 0, all faces converge to the "mean" face of CelebA. This is similar to figure (8) from [3]

## V. Discussion and Future Work

StyleGAN is currently the state-of-the-art in generating images, especially in the task of realistic face generation. In the sections above the procedure of re-training it on several public face datasets is discussed. The network was re-trained on two large publicly available datasets, CelebA and CASIA-WebFace. The Original StyleGAN has been trained on the FFHQ and CelebA-HQ face datasets which are of high quality and high resolution in [3]. As in this work, the training is performed with images of low resolution and lower quality, no comparison is being performed between the StyleGAN models of this work and the models from [3]. We trained StyleGAN models on different face datasets with different resolutions providing a useful tool for researchers, as to make these models available in [1]. Furthermore, it gives the opportunity to examine several aspects of StyleGAN. As mentioned StyleGAN is being trained on other large face datasets and further results will be presented at the ISSC conference later this year. It should be noted that this work is a first step and an important tool that will be used in order to understand how the size and quality of the original dataset affect the quality and distribution of the output data samples. Also, it will help to study how these tools could be used in order to build large, scalable datasets of synthetic facial data. Future works include, a study regarding the amount of data and variation needed in order to train StyleGAN effectively and a study regarding the relationship between the original samples used for training the StyleGAN models and the generated samples will be performed as well as examining the relationship between the generated samples.

## Acknowledgment

This research is funded under the SFI Strategic Partnership Program by Science Foundation Ireland (SFI) and FotoNation Ltd. Project ID:13/SPP/I2868 on Next Generation Imaging for Smartphone and Embedded.